%% file: acl2021.tex
\documentclass[11pt,a4paper]{article}
\usepackage[hyperref]{acl2021}
\usepackage{times}
\usepackage{latexsym}
\usepackage{amssymb}
\usepackage{amsmath} 
\usepackage{booktabs}
\usepackage{enumerate}
\usepackage{graphicx}
\usepackage{subfigure}
\usepackage{xspace}
\usepackage{float}
\usepackage{bbm}
\usepackage{bm}
\usepackage{multirow}
\usepackage{booktabs}
\usepackage{color}
\usepackage{framed}
\usepackage{stfloats}
\usepackage{iitem}
\usepackage[T1]{fontenc}
\usepackage{url}
\newcommand{\paratitle}[1]{\vspace{1.5ex}\noindent\textbf{#1}}
\newcommand{\ie}{\emph{i.e.,}\xspace}

\newcommand{\eg}{\emph{e.g.,}\xspace}

\newcommand{\ignore}[1]{}
\newcommand{\tabincell}[2]{\begin{tabular}{@{}#1@{}}#2\end{tabular}}

\usepackage{xcolor}
\definecolor{tOrange}{RGB}{255,165,0}
\definecolor{tBlue}{RGB}{24,116,205}
\definecolor{tPink}{RGB}{255,20,147}
\definecolor{tGreen}{RGB}{50,205,50}
\definecolor{tGold}{RGB}{255,215,0}

\usepackage{microtype}

\aclfinalcopy 

\title{Few-shot Knowledge Graph-to-Text Generation \\with Pretrained Language Models}

\author{
	Junyi Li\textsuperscript{\rm{1,3}}, 
	Tianyi Tang\textsuperscript{\rm{1}}, 
	Wayne Xin Zhao\textsuperscript{\rm{1,3,5}\thanks{\ \ Corresponding author}\ },
	Zhicheng Wei\textsuperscript{\rm{4}}, \\
	\textbf{Nicholas Jing Yuan}\textsuperscript{\rm{4}}
	{\rm and} \textbf{Ji-Rong Wen}\textsuperscript{\rm{1,2,3}} \\
	\textsuperscript{1}Gaoling School of Artificial Intelligence, Renmin University of China \\
	\textsuperscript{2}School of Information, Renmin University of China \\
	\textsuperscript{3}Beijing Key Laboratory of Big Data Management and Analysis Methods \\
	\textsuperscript{4}Huawei Cloud \\
	\textsuperscript{5}Beijing Academy of Artificial Intelligence, Beijing, 100084, China \\
	\texttt{\{lijunyi,jrwen,steven\_tang\}@ruc.edu.cn} \\ 
	\texttt{\{batmanfly,nicholas.jing.yuan\}@gmail.com} \quad \texttt{weizhicheng1@huawei.com}\\
}

\date{}

\begin{document}
\maketitle
\begin{abstract}
This paper studies how to automatically generate a natural language text that describes the facts in knowledge graph (KG). Considering the few-shot setting, we leverage the excellent capacities of  pretrained language models (PLMs) in language understanding and generation. We make three major technical contributions, namely representation alignment for bridging the semantic gap between KG encodings and PLMs, relation-biased KG linearization for deriving better input representations, and  multi-task learning for learning the correspondence between KG and text. Extensive experiments on three benchmark datasets have demonstrated the effectiveness of our model on KG-to-text generation task. 
	In particular, our model outperforms all comparison methods on both fully-supervised and few-shot settings. Our code and datasets are available at \url{https://github.com/RUCAIBox/Few-Shot-KG2Text}.
\end{abstract}

\input{sec-intro}

\input{sec-rel}
\input{sec-model}

\input{sec-exp}

\input{sec-con}
\input{sec-ack}

\bibliographystyle{acl_natbib}
\bibliography{graph2text}


\end{document}

%% file: sec-intro.tex
\section{Introduction}

\begin{figure}[tb]
	\centering
	\includegraphics[width=0.47\textwidth]{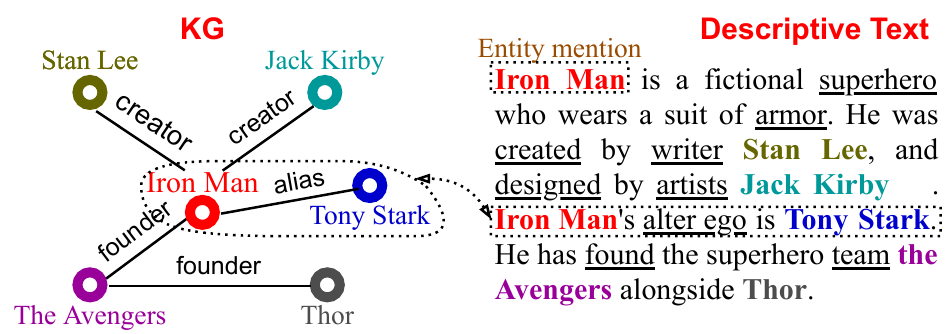}
	\caption{A knowledge graph (subgraph) with its descriptive text. The underlined words represent the context keywords about entities.}
	\label{fig-example}
\end{figure}

Knowledge graphs (KGs), such as Wikidata and DBpedia, are essential for many natural language processing (NLP) applications~\cite{abs-2002-00388}. To understand the structured information in KG, the task of KG-to-text generation has been proposed to automatically generate a descriptive text for a given knowledge graph~\cite{Koncel-Kedziorski19,RibeiroZGG20}. Figure~\ref{fig-example} illustrates a KG with the corresponding descriptive text, in which the nodes (\eg \emph{Stan Lee} and \emph{Iron Man}) represent entities and the edges (\eg \emph{creator} and \emph{alias}) describe the relations between connected entities.

In recent years, with the help of crowdsourcing platforms and information extraction (IE) systems, large-scale labelled pairs of KG and its descriptive text have been created, such as WikiBio~\cite{LebretGA16} and WebNLG Challenge~\cite{GardentSNP17}. Based on these datasets, data-driven models have shown impressive capabilities to produce informative and fluent text for a given KG~\cite{logan2019barack,MoryossefGD19}. However, due to the great expense in annotation process, it is not always feasible to generate large-scale labelled datasets for a variety of domains in practice. Motivated by this, we propose to study the task of \emph{few-shot KG-to-text generation} that aims to produce satisfactory output text given only a handful of (several hundred) labelled instances.   

To fulfil this task, we need to fully understand the  complicated semantic relations between entities from various domains, which is challenging with limited labelled data. 
Our solution is inspired by the excellent few-shot capabilities of pretrained language models (PLMs) on language understanding and generation tasks~\cite{BrownMRSKDNSSAA20,ChenECLW20,li2021textbox}. 
Pretrained on the large-scale corpora, PLMs encode vast amounts of world knowledge into their parameters~\cite{li2021survey}, which is potentially beneficial to understand and describe the KG facts in our task. 

However, applying PLMs to few-shot KG-to-text generation still faces two challenges. First, PLMs are usually pretrained on natural language text, while the KG inputs for our task are structured graphs. This semantic gap makes it difficult to effectively inject KG representations into PLMs especially with limited labelled instances. Second, unlike many other text generation tasks, KG-to-text generation requires faithful generation based on the understanding of KG facts. It needs to learn an accurate semantic correspondence between input KG and output text, which will be more difficult in few-shot settings.

To address the above issues, in this paper, we propose a few-shot KG-to-text generation model based on PLMs. There are three major technical contributions in our model. First, 
in order to bridge the semantic gap, we enforce the representation alignment by 
learning the correspondence between KG representations (encoded by graph neural networks) and PLM-based entity representations. Second, to feed KG into PLMs, we 
propose a relation-biased breadth-first search (RBFS) strategy to linearize KG into a well-planned entity sequence. Finally, we jointly train the primary text generation task and an auxiliary KG reconstruction task under the framework of multi-task learning.
This step further enhances the semantic correspondence between input KG and output text, based on which our model can generate faithful text about KG. 


To the best of our knowledge, we are the first study to investigate PLMs for few-shot KG-to-text generation. Extensive experiments on three benchmark datasets demonstrate the effectiveness of our few-shot KG-to-text generation model.

%% file: sec-rel.tex
\section{Related Work}

In this work, we mainly focus on generating text from knowledge graphs using PLMs.

\paratitle{KG-to-Text Generation}. Early works mainly centered around statistical methods, applying grammar rules to generate text \cite{KonstasL13, FlaniganDSC16}. Recently, neural based approaches have been proposed to generate text from linearized KG triples \cite{GardentSNP17}, however, unable to model structural information about KG. Many works explored how to encode the graph structure using Graph Neural Networks (GNNs) or Transformers explicitly. \citet{Koncel-Kedziorski19} leveraged a graph Transformer encoder to compute node representations by attending over local neighborhoods via self-attention. In contrast, \citet{RibeiroZGG20} focused on combining global and local message passing mechanisms based on GNNs, capturing complementary graph contexts. \citet{guo2020cyclegt} presented an unsupervised training method that can iteratively back translate between the text and graph data. Different from them, we explore how to utilize large PLMs for few-shot KG-to-text generation. 

\paratitle{Pretrained Language Model}. Recent years have witnessed prominent achievement of PLMs in NLP tasks~\cite{DevlinCLT19, radford2019language}. Pretrained on massive corpora, pretrained models showcase unprecedented generalization ability to solve related downstream tasks~\cite{li2021survey}. However, most of existing PLMs were conditioned on text data~\cite{radford2019language, LewisLGGMLSZ20}, lacking consideration of structured data input. \citet{ribeiro2020investigating} proposed to utilize PLMs for KG-to-text generation by randomly linearizing graph into a sequence of triples. While, these methods do not explicitly model the structural relations of KG, which is critical for generating faithful text. Our work aims to consider the KG structure and bridge the semantic gap between KG encodings and PLMs.

%% file: sec-model.tex
\section{Problem Formulation}
\label{sec-formulation}

KG-to-text generation~\cite{RibeiroZGG20} aims to automatically generate a natural language text that describes the facts in KG. 

Formally, the input KG consists of a set of triples, denoted as $\mathcal{G}=\{\langle e, r, e' \rangle | e, e' \in \mathcal{E}, r \in \mathcal{R}\}$, where $\mathcal{E}$ and $\mathcal{R}$ denote the entity set and relation set, respectively. A triple $\langle e, r, e' \rangle$ denotes the fact that relation $r$ exists between head entity $e$ and tail entity $e'$. Note that the input KG is a small and compact subgraph extracted from large-scale knowledge graphs (\eg DBpedia). Following~\citet{Koncel-Kedziorski19}, a text describing the input KG is usually available in this task. Let $\mathcal{V}$ denote the vocabulary. The target is to generate a natural language text $\mathcal{Y}=\langle w_1, ..., w_j, ..., w_T \rangle (w_j \in \mathcal{V})$ that represents the correct and concise semantics of entities and their relations in the given knowledge graph. The text contains a set of entity mentions $\mathcal{M}=\{m_e | m_e = \langle e, s_e, o_e \rangle, e \in \mathcal{E} \}$, where $e$ is the target entity, $s_e$ and $o_e$ are the start and end indices of this mention in text $\mathcal{Y}$, respectively. In other words, $\langle w_{s_e},...,w_{o_e} \rangle$ specially  corresponds to entity $e$. For entities with multiple mentions in text, we only keep the first mention of each entity in $\mathcal{M}$. By replacing each word of mentions with the token ``[MASK]'', we can obtain a masked text, denoted as $\mathcal{Y}_{[mask]}$, which is also taken as input for representation alignment in Section~\ref{sec-kg-repre}.

In practice, it is difficult to collect massive pairs of KG and its descriptive text for training. In this paper, we study the task of \emph{few-shot KG-to-text generation} with a handful of training instances (\eg 200 instances) based on a given PLM (\eg GPT-2).

\section{Approach}
\begin{figure}[!tb]
	\centering 
	\includegraphics[width=0.46\textwidth]{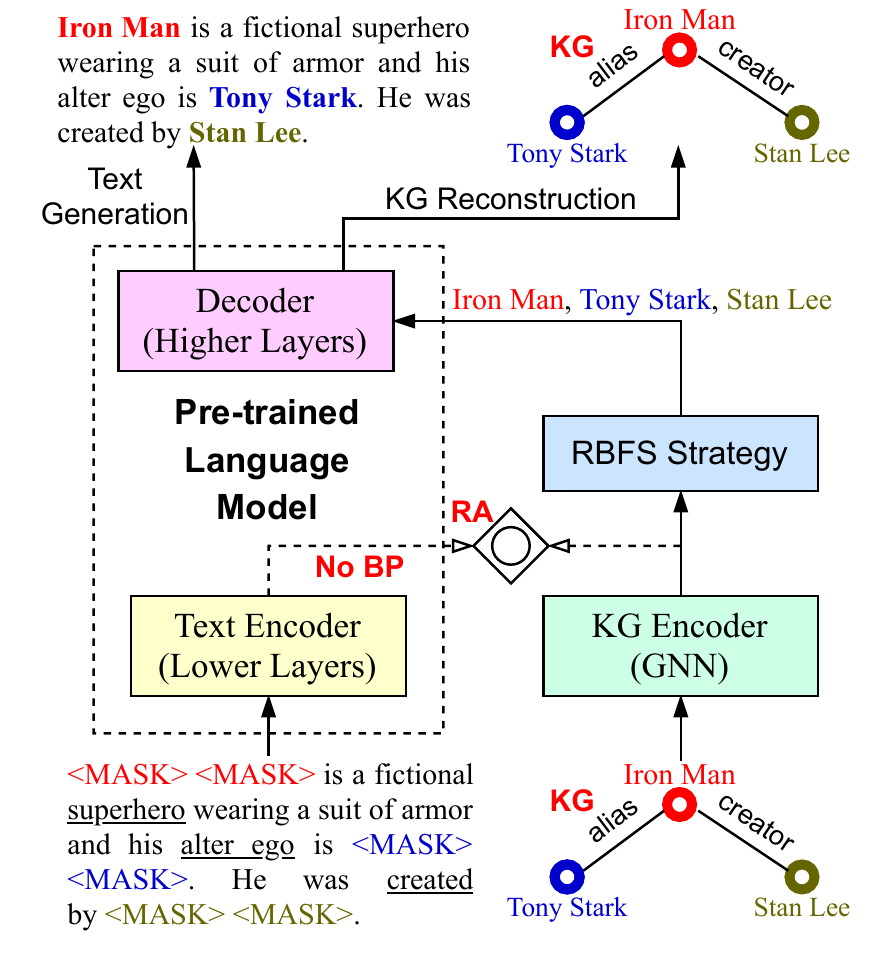} 
	\caption{Overview of our proposed model. ``RA'' and ``BP'' denote representation alignment and back propagation, respectively. We organize the PLM into lower layers and higher layers. The former provides PLM-based entity representations for alignment with KG encodings, and the latter acts as a decoder for generating text and reconstructing KG facts. After representation alignment, KG embeddings can be directly fed into the higher layers of PLMs for generating text.} 
	\label{fig-model} 
\end{figure}

For our task, two major challenges are how to learn effective input representations and capture the semantic correspondence between KG and text. To address the two challenges, we propose three major technical contributions, namely representation alignment between KG encodings and PLMs, relation-biased BFS strategy for KG linearization, and multi-task learning with KG reconstruction.  Figure~\ref{fig-model} presents an illustrative overview of our model. Next we will describe each part in detail.


\subsection{Representation Alignment}
\label{sec-kg-repre}

Unlike previous works~\cite{ribeiro2020investigating,yang2020improving} that directly transform KG into text sequence, we employ graph neural network (GNN) as knowledge graph encoder to explicitly encode entity relations in KG. Based on the input KG, GNN would produce a set of entity embeddings, which can be regarded as the input word embeddings of PLM for generating text. However, the GNN-based entity embeddings and the PLM-based word (entity) embeddings come from two distinct semantic spaces. 
To bridge such a semantic gap, we propose a representation alignment method to align the GNN-based and PLM-based entity embeddings in different semantic spaces. 

\paratitle{KG Encoder}. The GNN-based KG encoder aims to generate entity embeddings for KG. Let $\bm{v}_e \in \mathbb{R}^{d_E}$ denote the entity embedding for a general entity $e$ in KG, where $d_E$ is the embedding size. In our work, the entity embeddings are shared across different KGs and initialized with pretrained KG embeddings~\cite{YangYHGD14a}. We apply R-GCN~\cite{SchlichtkrullKB18} to generate entity embeddings by leveraging multi-relational information in KG. Then, the embedding of entity $e$ at the $l+1$-th layer of R-GCN can be computed as:
\begin{equation}\label{eq-RGCN}
\bm{v}^{(l+1)}_{e}= \sigma (\sum_{r \in \mathcal{R}} \sum_{e' \in \mathcal{N}^r_{e}} \bm{W}^{(l)}_r \bm{v}^{(l)}_{e'} + \bm{W}^{(l)}_0 \bm{v}^{(l )}_{e} ),
\end{equation}
where $\bm{W}^{(l)}_0$ and $\bm{W}^{(l )}_r$ are trainable matrices, and $\mathcal{N}^r_{e}=\{e' | \langle e, r, e' \rangle, \langle e', r, e \rangle \in \mathcal{G}\}$ denotes the set of neighbors of entity $e$ under relation $r$. Finally, after stacking $L$ times, the output entity embedding $\bm{v}^{(L)}_{e}$ from the last R-GCN layer is used as the final entity embedding $\bm{\tilde{v}}_{e}$. 

Note that, we represent an entity as a set of nodes. For instance, the entity \emph{Iron Man} in Figure~\ref{fig-example} will be represented by two nodes: one for the token \emph{Iron} and the other for the token \emph{Man}. This would  enhance the generalization ability of KG encoder on unseen entities, since it learns entity embeddings at the token level.

\paratitle{Text Encoder}. To obtain the PLM-based entity embeddings, we feed the masked text $\mathcal{Y}_{[mask]}$ into the text encoder, \ie the lower layers of PLM. As shown in Figure~\ref{fig-example}, compared with short entity mentions, the masked text contains rich context information about entities. Therefore, similar to masked language model~\cite{DevlinCLT19}, the embeddings of masked text can be computed as:
\begin{equation}\label{eq-mlm}
	\langle \bm{\hat{v}}_{w_1},...,\bm{\hat{v}}_{w_T} \rangle = \text{Lower-Layers}(\mathcal{Y}_{[mask]}),
\end{equation}
where the entity mention $m_e$ corresponds to the embedding sequence $\langle \bm{\hat{v}}_{w_{s_e}},...,\bm{\hat{v}}_{w_{o_e}} \rangle$ and the PLM-based entity embedding $\bm{\hat{v}}_e$ can be computed by an average pooling over this embedding sequence. 

To bridge the semantic gap, we model the representation alignment by minimizing the Euclidean distance in semantic space between the GNN-based and PLM-based entity embeddings as:
\begin{equation}\label{eq-ra-loss}
\mathcal{L}_{RA} = \sum_{e \in \mathcal{E}} \Vert \bm{\tilde{v}}_e - \bm{\hat{v}}_e \Vert_2,
\end{equation}
where $\bm{\tilde{v}}_e$ and $\bm{\hat{v}}_e$ are GNN-based and PLM-based entity embeddings, respectively. 

With representation alignment, the GNN-based entity embeddings can be aligned with the PLM-based entity embeddings in semantic space, which enables us to effectively inject KG representations into PLM for improving generation quality. 

\subsection{Knowledge Graph Linearization}
\label{sec-kg-linearization}

To feed the KG into decoder (\ie the higher layers of PLM), we need to linearize KG into an entity sequence. Previous work~\cite{yang2020improving,ribeiro2020investigating} usually relies on random or pre-defined rules, which is not flexible to model KG structures. Here, we propose to utilize breadth-first search (BFS) strategy to traverse KG. BFS, a graph traversal algorithm, starts at the root node and explores all the nodes at the present layer before moving on to the nodes at the next depth layer\footnote{https://en.wikipedia.org/wiki/Breadth-first\_search}. Here, we assume that nodes at the same layer potentially express relevant semantics and should be placed in close positions of the entity sequence. 

Furthermore, we observe that some relations are often lexicalized before others, \eg the nationality of a person often precedes the birthplace in descriptive text. Considering such relation priority, in this paper, we propose a \emph{relation-biased breadth first search (RBFS)} strategy to traverse and linearize KG into entity sequence. Specifically, we first compute RBFS weights $\alpha_{e'}$ for each entity $e'$ based on their relations as:
\begin{equation}
	\alpha_{e'} = \sigma(\bm{\tilde{v}}^\top_e \bm{W}_r^{(L)} \bm{\tilde{v}}_{e'}),~\langle e, r, e' \rangle \in \mathcal{G},
\end{equation}
where $\bm{W}_r^{(L)}$ is a relation matrix from Eq.~\ref{eq-RGCN}. Then, for two sibling entities $e'$ and $e''$ at the same layer, we traverse $e'$ before $e''$ if $\alpha_{e'}$ is greater than $\alpha_{e''}$, and vice versa. Finally, through RBFS, we can obtain a linearized entity sequence taken as input of the decoder for text generation. 

\subsection{KG-enhanced Multi-task Learning}
\label{sec-multi-task}

After obtaining the linearized entity sequence, we next take it as input and perform text generation. Different from other text generation tasks, KG-to-text generation aims to generate text reflecting the concise facts in KG. Inspired by~\citet{LiuLXMCS19}, we incorporate an auxiliary KG reconstruction task to reconstruct the facts in KG for learning the semantic correspondence between text and KG.

\paratitle{Text Generation}. The text generation task is performed upon the  higher layers of PLM. The objective is to maximize the likelihood of the reference text, which is equivalent to minimize the negative log-likelihood as:
\begin{equation}\label{eq-lm-loss}
	\mathcal{L}_{LM} = - \sum_{j=1}^{T} \log p_{gen}(w_j|w_1,...,w_{j-1};\mathcal{G}),
\end{equation}
where $p_{gen}$ is the generative probability from PLM. Besides, in KG-to-text generation, some tokens in descriptive text correspond to KG entities shown in Figure~\ref{fig-example}. The ability to copy entities from KG would enrich the generated text content, which can be achieved by the pointer generator~\cite{SeeLM17}. By feeding the hidden states of PLM and the token embedding, the copy probability $p^j_{copy}$ of the $j$-th token $w_j$ can be computed as: 
\begin{equation}
	p^j_{copy} = \sigma(\bm{W}_1 \bm{s}_j + \bm{W}_2 \bm{v}_{w_j} + b_{copy}),
\end{equation}
where $\bm{W}_1$, $\bm{W}_2$, and $b_{copy}$ are trainable parameters, $\bm{v}_{w_j}$ is the embedding of token $w_j$, and $\bm{s}_j$ is the $j$-th hidden state from the top layer of PLM. Then, we explicitly ``teach'' our model how to switch between generation and copy via the copy loss as:
\begin{equation}\label{eq-copy-loss}
	\mathcal{L}_{PG} = \sum_{w_j} p^j_{copy} + \sum_{w_k} (1 - p^k_{copy}).
\end{equation}
Our intuition is aimed at minimizing the copy probability $p^j_{copy}$ of token $w_j$ (generated from vocabulary) and maximizing the copy probability $p^k_{copy}$ of token $w_k$ (copied from KG entities). 

\paratitle{KG Reconstruction}. Following~\citet{SongWSZXGY20}, we formalize the KG reconstruction task as predicting the relations between any two entities. In detail, given a head entity $e$ and a tail entity $e'$ in generated text, we can obtain the hidden states of their mentions from the top layer of decoder, \ie $\langle \bm{s}_{s_e},...,\bm{s}_{o_e} \rangle$ and $\langle \bm{s}_{s_{e'}},...,\bm{s}_{o_{e'}} \rangle$. Then, the entity hidden states $\bm{h}_e$ and $\bm{t}_{e'}$ can be computed by an average pooling over their mention hidden states. The probability for a relation $r$ is calculated as:
\begin{equation}
	p(r|e,e') = \text{softmax}(\bm{W}_3[\bm{h}_e;\bm{t}_{e'};\bm{h}_e \odot \bm{t}_{e'}] + \bm{b}_2),
\end{equation}
where $\bm{W}_3$ and $\bm{b}_2$ are trainable parameters. The loss for reconstructing KG is also defined as the negative log-likelihood of all target triples in KG:
\begin{equation}\label{eq-gr-loss}
	\mathcal{L}_{GR} = -\sum_{\langle e,r,e' \rangle \in \mathcal{G}} \log p(r|e,e').
\end{equation}

By incorporating the KG reconstruction task, our model is able to capture the semantic correspondence between input KG and output text, which further improves generating faithful text.

Finally, the total training loss consists of text generation loss $\mathcal{L}_{LM}$ (Eq.~\ref{eq-lm-loss}), copy loss $\mathcal{L}_{PG}$ (Eq.~\ref{eq-copy-loss}), representation alignment loss $\mathcal{L}_{RA}$ (Eq.~\ref{eq-ra-loss}) and KG reconstruction loss $\mathcal{L}_{GR}$ (Eq.~\ref{eq-gr-loss}) as:
\begin{equation}\label{eq-total-loss}
	\mathcal{L}_{total} = \mathcal{L}_{LM} + \lambda_1 \mathcal{L}_{PG} + \lambda_2 \mathcal{L}_{RA} + \lambda_3 \mathcal{L}_{GR},
\end{equation}
where $\lambda_1$, $\lambda_2$ and $\lambda_3$ are combination coefficients.

\subsection{Discussion and Learning}
In this part, we present the model discussion and the model optimization.

\paratitle{Few-shot Learning}. In few-shot KG-to-text generation, the key lies in how to bridge the semantic gap between KG and PLMs with limited dataset. To achieve this goal, we first utilize representation alignment in Section~\ref{sec-kg-repre} to align the semantic space between KG encodings and PLMs, and then introduce a KG reconstruction task in Section~\ref{sec-multi-task} to further learn the semantic correspondence between input KG and output text. Besides, we observe that KG entities are often multi-word expressions. To deal with unseen entities in few-shot learning, we employ the Byte Pair Encoding (BPE)~\cite{SennrichHB16a} and sub-word vocabulary~\cite{radford2019language} to split entity words into smaller semantic units. Our work is also empowered by the excellent few-shot capacities of PLMs with vast amounts of world knowledge learned from large-scale corpora. 

\paratitle{Optimization}. For PLM, we employ BART-Large model~\cite{LewisLGGMLSZ20}. Specially, we adopt the first 6 layers of BART encoder as the lower layers, and the remaining 6 layers of BART encoder and BART decoder as the higher layers. Note that, the target text and text encoder will not be used at test time. In particular, the target text is just used at training time and encoded as PLM-based entity embeddings for representation alignment, while the alignment is not needed at test time. We optimize all parameters according to the total loss in Eq.~\ref{eq-total-loss} with the OpenAI AdamW optimizer~\cite{LoshchilovH19}. The learning rate, batch size, R-GCN layers and embedding size are set to 1e-5, 20, 2 and 1024, respectively. The weights $\lambda_1$, $\lambda_2$ and $\lambda_3$ in Eq.~\ref{eq-total-loss} are set to 0.7, 0.5 and 0.5, respectively, according to performance on validation set. During inference, we apply the beam search method with a beam size of 8.

%% file: sec-exp.tex
\section{Experiments}

In this section, we first set up the experiments, and then report the
results and analysis.


\subsection{Experimental Setup}
\paratitle{Datasets}. To evaluate our model on few-shot KG-to-text generation, we conduct experiments on three benchmarks, including AGENDA~\cite{Koncel-Kedziorski19}, WebNLG~\cite{GardentSNP17} and GenWiki Fine~\cite{JinGQZ20}. We adopt three large domains (\ie \textit{Airport}, \textit{Building} and \textit{Food}) for WebNLG and two large domains (\ie \textit{Sports} and \textit{Games}) for GenWiki. Table~\ref{tab-data} shows the statistics for each dataset. Each instance of these datasets contains a knowledge graph in the form of triples and a target text describing the graph. The three datasets have originally provided the alignment records from entity mentions to KG entities. Take an example from WebNLG dataset ``AGENT-1 is located in PATIENT-1'': the entity mention is tagged as ``AGENT-1'' and the tag ``AGENT-1'' maps to the entity ``11th\_Mississippi\_Infantry\_Monument'' in KG. If such alignments are not available, we can utilize entity linking tools (\eg NER packages) for preprocessing. 

\begin{table}[tbp]
	\centering
	\small
	\begin{tabular}{l|r r r r}
		\toprule[1pt]
		\textbf{Dataset} & \#Train & \#Valid & \#Test & \#Relations \\
		\midrule[0.7pt]
		\textbf{AGENDA} & 29,720 & 1,000 & 10,000 & 42 \\
		\textbf{WebNLG} & 7,362 & 1,389 & 5,427 & 107 \\
		\textbf{GenWiki}	 & 48,020 & 1,000 & 10,000 & 250 \\
		\bottomrule[1pt]
	\end{tabular}%
	\caption{Statistics of three datasets.}
	\label{tab-data}%
\end{table}

\paratitle{Baselines}. We make a comparison against five KG-to-text generation models:

\textbullet~\underline{\emph{GraphWriter}}~\citep{Koncel-Kedziorski19} introduces a graph transformer encoder and a sequence decoder for generating text based on KG.

\textbullet~\underline{\emph{CGE-LW}}~\citep{RibeiroZGG20} proposes a graph-to-text model by  combining both global and local node aggregation strategies.

\textbullet~\underline{\emph{CycleGT}}~\citep{guo2020cyclegt} jointly learns two dual tasks (graph-to-text generation and text-to-graph relation classification) via cycle training.

\textbullet~\underline{\emph{BART-Base/Large}}~\citep{ribeiro2020investigating} linearizes the KG into sequence and applies BART-Base/Large~\cite{LewisLGGMLSZ20} to generate text.

\textbullet~\underline{\emph{T5-Base/Large}}~\citep{ribeiro2020investigating} linearizes KG into a triple sequence and employs T5-Base/Large~\cite{raffel2020exploring} to generate text.

\begin{table*}[t]
\renewcommand\arraystretch{1.1}
\begin{minipage}[t]{\linewidth}
  \centering
     \makeatletter\def\@captype{table}\makeatother\small
     \begin{tabular}{l c c c c c c c c c c c c}
			\toprule[1pt]
			\textbf{Datasets} & \multicolumn{4}{c}{\textsc{AGENDA}} & \multicolumn{4}{c}{\textsc{WebNLG}} & \multicolumn{4}{c}{\textsc{GenWiki Fine}} \\
			\cmidrule(r){1-1}\cmidrule(r){2-5}\cmidrule(r){6-9}\cmidrule(r){10-13}
			\textbf{\#Metrics} & B-4 & R-L & CIDEr & Chrf & B-4 & R-L & CIDEr & Chrf & B-4 & R-L & CIDEr & Chrf \\
			\midrule[0.5pt]
			GraphWriter & 15.30 & 22.03 & 0.24 & 38.33 & 45.84 & 60.62 & 3.14 & 55.53 & 29.73 & 55.46 & 2.68 & 46.87\\
			CGE-LW & 18.01 & 25.62 & 0.33 & 46.69 & 48.60 & 62.52 & 3.85 & 58.66 & 30.67 & 56.37 & 3.20 & 47.79 \\
			CycleGT & 20.16 & 25.77 & 0.69 & 48.26 & 50.20 & \underline{68.30} & 3.81 & 68.91 & 38.57 & 59.37 & 3.50 & 62.46 \\
			BART-base & 22.01 & 26.44 & 0.90 & 48.02 & 49.81 & 63.10 & 3.45 & 67.65 & 48.20 & 59.21 & 4.02 & 65.80\\
			BART-large & \underline{23.65} & 28.76 & \underline{1.15} & \underline{50.44} & 52.49 & 65.61 & 3.50 & 72.00 & \textbf{50.70} & \underline{61.90} & \underline{4.51} & \underline{68.15} \\
			T5-base & 20.59 & 29.41 & 0.81 & 48.15 & 48.86 & 65.57 & 3.99 & 66.08 & 45.72 & 58.28 & 3.74 & 65.68 \\
			T5-large & 22.15 & \underline{30.68} & 0.87 & 48.88  & \underline{58.78} & 68.22 & \underline{4.10} & \underline{74.40} & 47.11 & 60.64 & 3.74 & \textbf{68.47} \\
			\cline{1-13}
			Ours & \textbf{25.15} & \textbf{35.12} & \textbf{3.23} & \textbf{55.89} & \textbf{61.88} & \textbf{75.74} & \textbf{6.03} & \textbf{79.10} & \underline{48.46} & \textbf{65.65} & \textbf{5.19} & 64.00 \\
			\bottomrule[1pt]
		\end{tabular}
		\caption{Performance comparisons of different methods for fully-supervised KG-to-text generation under three domains. B-$n$ and R-$n$ are short for BLEU-$n$ and ROUGE-$n$. \textbf{Bold} and \underline{underline} fonts denote the best and the second best methods (the same as below).\\}
		\label{tab:main-results}
  \end{minipage}
  \begin{minipage}[t]{\linewidth}
   \centering
        \makeatletter\def\@captype{table}\makeatother\small
        \begin{tabular}{l r r r r r r r r r r r r}
			\toprule[1pt]
			\textbf{Datasets} & \multicolumn{4}{c}{\textsc{AGENDA}} & \multicolumn{4}{c}{\textsc{WebNLG}} & \multicolumn{4}{c}{\textsc{GenWiki Fine}} \\
			\cmidrule(r){1-1}\cmidrule(r){2-5}\cmidrule(r){6-9}\cmidrule(r){10-13}
			\textbf{\#Instances} & 50 & 100 & 200 & 500 & 50 & 100 & 200 & 500 & 50 & 100 & 200 & 500 \\
			\midrule[0.5pt]
			BART-large & 5.71 & 6.15 & 7.59 & 10.71 & 9.05 & 15.70 & 19.38 & 27.91 & 9.14 & 13.38 & 15.39 & 24.14 \\
			T5-large & 2.69 & 2.73 & 4.65 & 7.52 & 7.18 & 14.52 & 16.88 & 21.68 & 6.30 & 6.36 & 10.37 & 17.72 \\
			\cline{1-13}
			Ours & \textbf{6.22} & \textbf{9.40} & \textbf{10.21} & \textbf{17.93} & \textbf{10.60} & \textbf{17.46} & \textbf{20.00} & \textbf{31.79} & \textbf{10.75} & \textbf{14.44} & \textbf{16.84} & \textbf{28.89} \\ 
			\bottomrule[1pt]
		\end{tabular}
		\caption{BLEU-4 results of different methods for few-shot KG-to-text generation under three domains. To mitigate the randomized effects of samples, we report the average results over five training runs (the same as below).\\}
		\label{tab:fs-results-b4}
   \end{minipage}
   \begin{minipage}[t]{\linewidth}
   \centering
        \makeatletter\def\@captype{table}\makeatother\small
   		\begin{tabular}{l r r r r r r r r r r r r}
			\toprule[1pt]
			\textbf{Datasets} & \multicolumn{4}{c}{\textsc{AGENDA}} & \multicolumn{4}{c}{\textsc{WebNLG}} & \multicolumn{4}{c}{\textsc{GenWiki Fine}} \\
			\cmidrule(r){1-1}\cmidrule(r){2-5}\cmidrule(r){6-9}\cmidrule(r){10-13}
			\textbf{\#Instances} & 50 & 100 & 200 & 500 & 50 & 100 & 200 & 500 & 50 & 100 & 200 & 500 \\
			\midrule[0.5pt]
			BART-large & 14.33 & 15.28 & 16.94 & 20.70 & 22.57 & 26.21 & 30.68 & 49.34 & 26.59 & 29.60 & 34.56 & 47.50 \\
			T5-large & 14.11 & 14.17 & 15.88 & 21.72 & 20.80 & 22.71 & 24.18 & 38.36 & 21.02 & 21.36 & 20.07 & 35.72 \\
			\cline{1-13}
			Ours & \textbf{15.10} & \textbf{16.65} & \textbf{18.88} & \textbf{25.72} & \textbf{24.80} & \textbf{28.38} & \textbf{33.12} & \textbf{55.13} & \textbf{28.02} & \textbf{31.36} & \textbf{38.07} & \textbf{50.72} \\
			\bottomrule[1pt]
		\end{tabular}
		\caption{ROUGE-L results of different methods for few-shot KG-to-text generation under three domains.}
		\label{tab:fs-results-rl}
\end{minipage}
\end{table*}

Among these baselines, \emph{GraphWriter} and \emph{CGE-LW} are GNN-based generation models; \emph{CycleGT} is an unsupervised model using cycle training; \emph{GPT2-Base/Large} and \emph{BART-Base/Large} are the most relevant comparisons, which also employ PLMs in KG-to-text generation. These baselines were trained on the whole training dataset, \ie all KG-text pairs. Following previous few-shot work~\cite{ChenECLW20}, we train our model on different few-shot settings with training dataset size ranging from 50, 100, 200 to 500. All the comparison methods are optimized based on validation performance. In our model, the entity embeddings of GNN are initialized with pretrained KG embeddings and the GNN weights are transferred from CGE-LW. We also pretrain GNN weights based on the large-scale KG, \ie Wikipedia. Based on the pretrained entity embeddings and weights, we continue to train our model.

\paratitle{Evaluation Metrics}. For performance comparison, we adopt five automatic evaluation metrics widely used by previous graph-to-text work~\cite{guo2020cyclegt}, \ie BLEU~\cite{papineni2002bleu},  ROUGE~\cite{lin2004rouge}, CIDEr~\cite{VedantamZP15} and CHRF++~\cite{Popovic15}. Specifically, BLEU-$n$ and ROUGE-$n$ compute the ratios of overlapping $n$-grams between generated and real text, CIDEr computes the TF-IDF weights for each $n$-gram in generated/real text, and CHRF++ computes  F-score averaged on both character- and word-level $n$-grams. 

\subsection{Main Results}


Table~\ref{tab:main-results},~\ref{tab:fs-results-b4}, and~\ref{tab:fs-results-rl} present the fully-supervised and few-shot results of our model and other baselines, respectively. 

First, by combining global and local entity context, CGE-LW performs better than GraphWriter. Furthermore, with two elaborate designed dual tasks, CycleGT becomes the best non-PLM baseline, outperforming GraphWriter and CGE-LW. 

Second, as the most direct comparison with our model, BART-Base/Large and T5-Base/Large perform better than baselines by leveraging encoded semantics in PLMs, which reveals the feasibility of utilizing PLMs for KG-to-text generation. 

Finally, we observe that our model achieves the best performance on both fully-supervised and few-shot settings. Large-scale PLMs can encode world knowledge by reading a large amount of text, making it easier to recover KG facts. Given only a handful of examples, the performances of baselines drop drastically, while the performance of our model only descents slightly. Furthermore, with only 500 labelled instances, our model improves over CGE-LW and CycleGT, and achieves the best performance in most cases. Compared to these PLM-based KG-to-text baselines, we adopt GNN to explicitly encode KG structure and representation alignment to bridge the semantic gap between PLM and GNN. This helps produce effective semantic representations for few-shot learning. Furthermore, we incorporate an auxiliary KG reconstruction task to learn semantic correspondence between input KGs and output text. These results indicate that our model can achieve more superior performance on KG-to-text generation task in a few-shot setting. 

\begin{table}[t]
	\small
	\centering
	\begin{tabular}{ l c c c c}
		\toprule[1pt]
		\textbf{Models} & \textbf{B-4} & \textbf{R-L} & \textbf{CIDEr} & \textbf{Chrf} \\
		\midrule[0.7pt]
		Ours & \textbf{31.79} & \textbf{55.13} & \textbf{3.94} & \textbf{57.38} \\
		\midrule[0.7pt]
		w/o RA & 23.14 & 41.34 & 1.90 & 43.34 \\
		w/o GR & 27.56 & 46.69 & 2.82 & 48.90 \\
		w/o PG & 29.30 & 48.66 & 3.58 & 53.44 \\
		\bottomrule[1pt]
	\end{tabular}
	\caption{Ablation analysis on \textsc{WebNLG} dataset.} 
	\label{tab:ablation-results}
\end{table}

\subsection{Detailed Analysis}
Next, we conduct detailed analysis experiments on our model.
We only report the test results on \textsc{WebNLG} dataset with 500 training instances due to similar findings in other datasets.

\paratitle{Ablation Analysis}. In our ablation study, we evaluate the effect of each loss $\mathcal{L}_{PG}$, $\mathcal{L}_{RA}$ and $\mathcal{L}_{GR}$ on the overall model performance. Here, we consider three variants: 

\textbullet~\emph{w/o PG}: the variant removes the copy loss $\mathcal{L}_{PG}$. 

\textbullet~\emph{w/o RA}: the variant removes the representation alignment loss $\mathcal{L}_{RA}$. 

\textbullet~\emph{w/o GR}: the variant removes the KG reconstruction loss $\mathcal{L}_{GR}$. 

As can be seen from Table~\ref{tab:ablation-results}, by removing any of the three losses, the BLEU/ROUGE/CIDEr performance drops compared to the complete model, especially removing $\mathcal{L}_{RA}$ and $\mathcal{L}_{GR}$. The proposed representation alignment bridges the semantic gap between PLM and GNN, which is helpful for adapting KG representations to PLM. The KG reconstruction task learns the correspondence between KG and text ensuring faithful generation about KG. We also observe a small performance drop by removing $\mathcal{L}_{PG}$. It is likely because PLM has learned some common phrase expressions about these KG facts from large-scale pretraining corpus.

\begin{figure}[tb]
	\centering
	\includegraphics[width=0.43\textwidth]{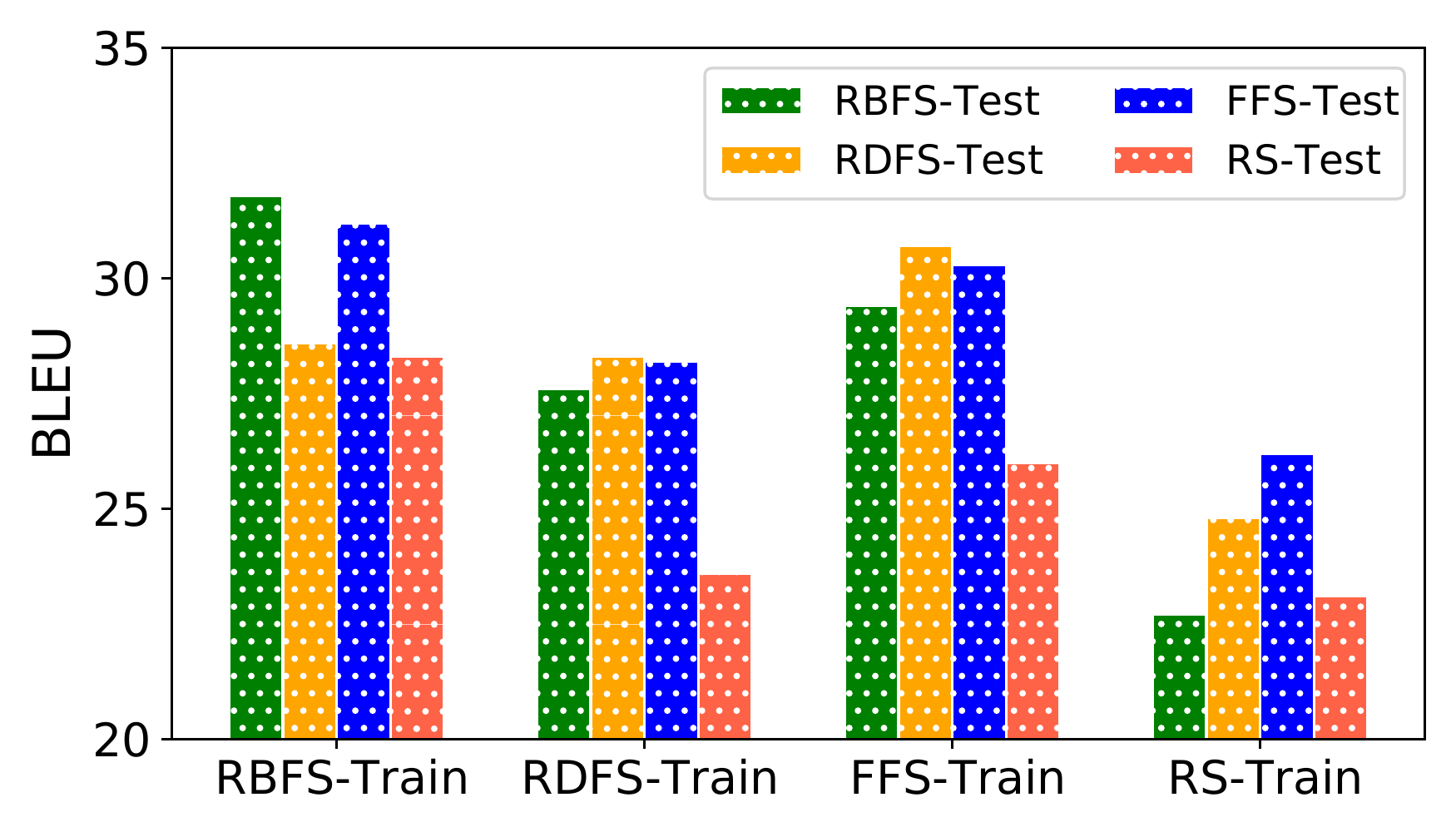}
	\caption{Linearization analysis on \textsc{WebNLG} dataset.}
	\label{fig-linearization-analysis}
\end{figure}

\begin{table}[t]
	\centering
	\small
	\begin{tabular}{l c c c }
		\toprule[1pt]
		\textbf{Models} & \textbf{\#Supp.}$\uparrow$ & \textbf{\#Cont.}$\downarrow$ &  \textbf{Naturalness}$\uparrow$ \\
		\midrule[0.7pt]
		Gold & 4.40 & 0.36 & 4.26 \\
		\hline
		Ours & \underline{3.77} & \underline{1.01} & \underline{3.96} \\
		BART-Large & 3.20 & 1.90 & 3.55 \\
		CEG-LW & 2.87 & 2.13 & 2.56 \\
		\bottomrule[1pt]
	\end{tabular}
	\caption{Human evaluation on  \textsc{WebNLG} dataset. Cohen’s kappa coefficients for labelling three factors are as follows: 0.78, 0.71, and 0.75.} 
	\label{tab:human-results}
\end{table}

\paratitle{KG Linearization Analysis}. In Section~\ref{sec-kg-linearization}, we propose a novel relation-biased BFS (RBFS) strategy to linearize the input KG into entity sequence. To verify the effectiveness of this strategy, we conduct linearization analysis by comparing RBFS with three traversal strategies, including relation-biased depth-first search (RDFS), forest fire search (FFS) and random search (RS). Specifically, RDFS combines both DFS and the relation factor similar to RBFS, where DFS starts at the root node and explores as far as possible along each branch before backtracking\footnote{https://en.wikipedia.org/wiki/Depth-first\_search}; FFS is a randomized version of RBFS  randomly exploring all the nodes at the same layer~\cite{LeskovecF06}; and RS  randomly traverses all the nodes in the input KG.
By re-training our model with the above three strategies, we report the comparison of BLEU results in Figure~\ref{fig-linearization-analysis}. It can be observed that, RBFS and FFS strategies achieve better results compared to the rest strategies. 
Nodes at the same layer tend to express more relevant semantics, thus searching by layer could produce more reasonable and coherent entity sequence especially considering the relations of entities as our RBFS strategy. 


\begin{table*}[t]
	\centering
	\small
	\begin{tabular}{c | c | c | c }
		\toprule[1pt]
		\multirow{2}[13]{*}{\textbf{Real}}
		&\tabincell{c}{Knowledge\\Graph} & \begin{minipage}{0.36\textwidth}\includegraphics[width=57mm, height=19.5mm]{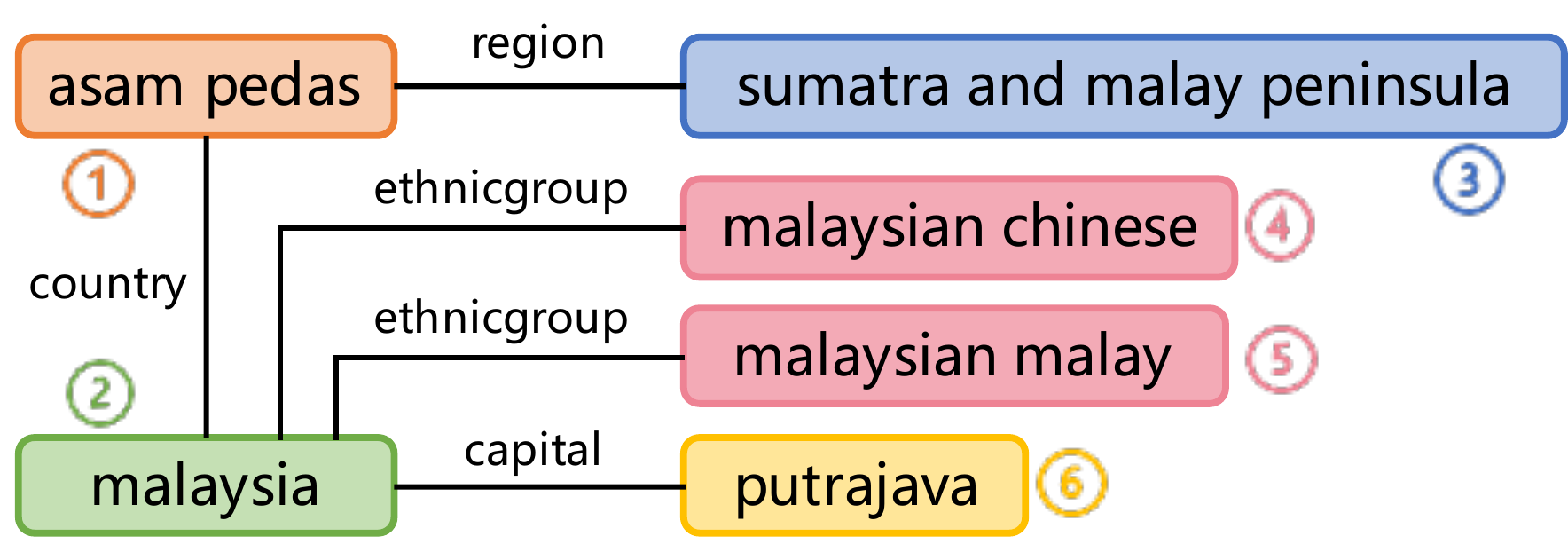}\end{minipage} & \begin{minipage}{0.38\textwidth}\includegraphics[width=56mm, height=19.5mm]{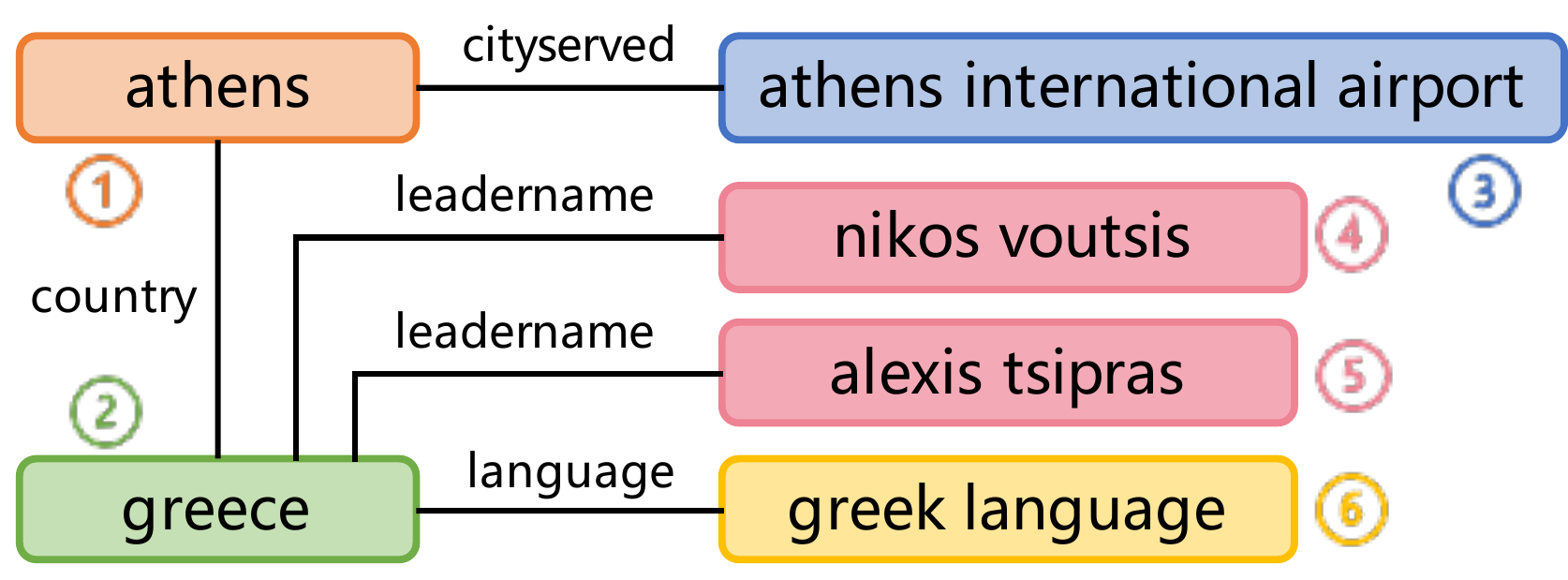}\end{minipage} \\
		\cline{2-4}
		&Reference & \tabincell{l}{\textbf{asam pedas} is a food found in the \underline{region} of \\\textbf{sumatra and malay peninsula} in \textbf{malaysia} , \\the \underline{capital} of which is \textbf{putrajaya} , and whose \\\underline{ethnic groups} include \textbf{malaysian malay} and \\\textbf{malaysian chinese} .} & \tabincell{l}{\textbf{athens international airport} \underline{serves} the \\ city \textbf{athens} in \textbf{greece} , \textbf{greek language} is \\\underline{spoken} in greece and the \underline{leaders} names in \\greece are \textbf{alexis tsipras} and \textbf{nikos voutsis} .} \\
		\midrule[0.7pt]
		\multirow{2}[9]{*}{\textbf{BART}}
		&\tabincell{c}{Linearized\\KG} & \textcolor{tOrange}{\textcircled{1}}\textcolor{tBlue}{\textcircled{3}}$\rightarrow$\textcolor{tOrange}{\textcircled{1}}\textcolor{tGreen}{\textcircled{2}}$\rightarrow$\textcolor{tOrange}{\textcircled{1}}\textcolor{tGold}{\textcircled{6}}$\rightarrow$\textcolor{tGreen}{\textcircled{2}}\textcolor{tPink}{\textcircled{5}}$\rightarrow$\textcolor{tGreen}{\textcircled{2}}\textcolor{tPink}{\textcircled{4}} & \textcolor{tOrange}{\textcircled{1}}\textcolor{tGreen}{\textcircled{2}}$\rightarrow$\textcolor{tGreen}{\textcircled{2}}\textcolor{tPink}{\textcircled{4}}$\rightarrow$\textcolor{tGreen}{\textcircled{2}}\textcolor{tPink}{\textcircled{5}}$\rightarrow$\textcolor{tOrange}{\textcircled{1}}\textcolor{tBlue}{\textcircled{3}}$\rightarrow$\textcolor{tGreen}{\textcircled{2}}\textcolor{tGold}{\textcircled{6}} \\
		\cline{2-4}
		&\tabincell{c}{Generated\\Text} & \tabincell{l}{\textbf{asam pedas} is a dish from \textbf{malaysia and} \\\textbf{sumatra} where the capital is \textbf{putrajava} . \\\textbf{malaysian malay} and \textbf{chinese} are ethnic \\groups in sumatra .} & \tabincell{l}{\textbf{athens} in \textbf{greece} is led by \textbf{alexis tsipras}  \\ and is served by \textbf{athens international} \\\textbf{airport} greece speaks \textbf{greek language} .} \\
		\midrule[0.7pt]
		\multirow{2}[10]{*}{\textbf{Ours}}
		&\tabincell{c}{Linearized\\KG} & \textcolor{tOrange}{\textcircled{1}}$\rightarrow$\textcolor{tBlue}{\textcircled{3}}$\rightarrow$\textcolor{tGreen}{\textcircled{2}}$\rightarrow$\textcolor{tGold}{\textcircled{6}}$\rightarrow$\textcolor{tPink}{\textcircled{5}}$\rightarrow$\textcolor{tPink}{\textcircled{4}} & \textcolor{tOrange}{\textcircled{1}}$\rightarrow$\textcolor{tBlue}{\textcircled{3}}$\rightarrow$\textcolor{tGreen}{\textcircled{2}}$\rightarrow$\textcolor{tGold}{\textcircled{6}}$\rightarrow$\textcolor{tPink}{\textcircled{5}}$\rightarrow$\textcolor{tPink}{\textcircled{4}}  \\
		\cline{2-4}
		&\tabincell{c}{Generated\\Text} & \tabincell{l}{\textbf{asam pedas} \underline{comes from} the \underline{region} of \textbf{sumatra} \\\textbf{and malay peninsula} in \textbf{malaysia} , where the \\\underline{capital} is \textbf{putrajava} , \textbf{malaysian malay} and \\\textbf{malaysian chinese} are \underline{ethnic groups} .} & \tabincell{l}{\textbf{athens} is \underline{served} by \textbf{athens international} \\\textbf{airport} in \textbf{greece} , which \underline{speaks} \textbf{greek} \\textbf{language} . greece is \underline{led} by \textbf{alexis} \\\textbf{tsipras} and \textbf{nikos voutsis} .}  \\
		\bottomrule[1pt]
	\end{tabular}
	\caption{Sample text generated by BART-Large baseline and our model from the \emph{Food} and \emph{Airport} domains of the \textsc{WebNLG} benchmark.  Since BART linearizes KG as triple sequence and an entity may involve in several triples, there are repeated entities used by BART (we omit the relations between entities). \textbf{Bold} and \underline{underlined} words correspond to entity words and keywords.} 
	\label{tab:qualitative}
\end{table*}

\paratitle{Human Evaluation}. Following previous work in data-to-text~\cite{ChenECLW20}, we conduct human evaluation on the generated text. We randomly sample 200 KG subgraphs along with corresponding generated text from CGE-LW, BART-Large and our model. In order to reduce the variance caused by human, three workers were asked to score the text with respect to two aspects: \textit{Factual correctness} and \textit{Language naturalness}. The first criterion evaluates how well the generated text correctly conveys information in the KG, by counting the number of facts in text supported by the KG (denoted as \#Supp.) and contradicting with or missing from the KG (denoted as \#Cont.). The second criterion evaluates whether the generated text is grammatically correct and fluent. The scoring mechanism adopts a 5-point Likert scale~\cite{likert1932technique}, ranging from 1-point~(``very terrible'') to 5-point~(``very satisfying''). We further average the three scores from the three human judges over the 200 inputs. The results in Table \ref{tab:human-results} show that our model produces more fidelity and fluent texts than previous models. In our approach, the KG reconstruction task and pointer generator enhance the awareness of KG facts and alleviate producing incorrect facts. Also, with some learned common phrase expressions in PLMs, our model can generate natural text while keeping fidelity.

\paratitle{Qualitative Analysis}. In this part, we present intuitive explanations why our model performs well. Table~\ref{tab:qualitative} presents two descriptions and the corresponding generated entity sequences and texts by BART-Large baseline and our model. As we can see, based on  KG linearization, the generated texts by our model show reasonable and similar content sketch with real texts (\eg \emph{peninsula (region)}$\rightarrow$\emph{malaysia (country)}$\rightarrow$\emph{putrajava (capital)}). Besides, the baseline model incorrectly merges and generates unfaithful facts (\eg \emph{malaysia and sumatra}) or misses facts (\eg \emph{nikos voutsis}), while our model describes all the KG facts correctly. This improvement could be attributed to the KG reconstruction task, which enables our model to learn the correspondence between the input KG facts and output text. Finally, the entity words in our generated text are enriched and connected by meaningful keywords (\eg entity \emph{greek language} and keyword \emph{speaks}). The reason might be that, with the help of representation alignment, the GNN entity embeddings are aligned with the PLM word embeddings.

%% file: sec-con.tex
\section{Conclusion}

This paper presented a few-shot KG-to-text generation model based on PLMs. We make three important technical contributions, namely representation alignment for bridging the semantic gap between KG encodings and PLMs, relation-biased KG linearization for deriving better input KG representations, and  multi-task learning for learning the correspondence between KG and text. 
Extensive experiments on three benchmark datasets demonstrate the effectiveness of our few-shot KG-to-text generation model. 
As future work, we will consider adopting KG-enhanced PLMs~\cite{ZhangHLJSL19,PetersNLSJSS19} for improving the task performance, which explicitly inject knowledge information into PLMs.


%% file: sec-ack.tex
\section*{Acknowledgement}
This work was partially supported by the National Natural Science Foundation of China under Grant No. 61872369 and 61832017, Beijing Academy of Artificial Intelligence (BAAI) under Grant No. BAAI2020ZJ0301, Beijing Outstanding Young Scientist Program under Grant No. BJJWZYJH012019100020098, the Fundamental Research Funds for the Central Universities, and the Research Funds of Renmin University of China under Grant No.18XNLG22 and 19XNQ047. Xin Zhao is the corresponding author.